\documentclass[11pt,fleqn,leqno]{article}

\usepackage[arrow,dvips,frame,matrix,ps]{xy}

\usepackage{ranlp97}

\usepackage{amsfonts}
\usepackage{booktabs}
\usepackage[hang]{caption}
\usepackage{avm}

\avmfont{\scriptsize\scshape}
\avmsortfont{\scriptsize\itshape}
\avmvalfont{\scriptsize\rm}
\avmhskip{0.5 em}
\avmvskip{0.05 ex}
\newcommand{\avmt}[1]{{\avmoptions{center}\begin{avm}#1\end{avm}}}
\newcommand{\avmvalue}[1]{{\avmjvalfont#1}}
\avmoptions{center}

\newcommand{\dq}{"}
\newcommand{\unif}{\mbox{\footnotesize$\, \sqcap \, $}}
\newcommand{\bigunif}{\mbox{\large$\, \sqcap \, $}}
\newcommand{\bfbigunif}{\mbox{\large$\, \sqcap\hspace*{-0.62 em}\sqcap \, $}}
\newcommand{\dunif}{\mbox{\footnotesize$\, \stackrel{\mbox{\tiny$<$}}{\sqcap} \, $}}

\newenvironment{cop}{%
  \list{}{\slshape%
    \leftmargin\parindent%
  }}{\endlist}
\newcommand{\coc}[1]{\textit{#1}}
\newcommand{\q}{\hspace*{1.0 em}}
\newcommand{\cow}[1]{\textsl{#1}}
\newcommand{\quw}[1]{\textit{#1}}
\newcommand{\emw}[1]{\textbf{#1}}
\newcommand{\eg}{e.\,g.,\ }
\newcommand{\ie}{i.\,e.,\ }
\newcommand{\car}{\makebox{\^{}}\-}

\title{Partial Evaluation for Efficient Access to Inheritance Lexicons}

\author{Sven Hartrumpf\thanks{\ \ Partial evaluation as described in this
paper has been successfully applied
to the lexicon in the \quw{Virtuelle Wissensfabrik}
(\quw{Virtual Knowledge Factory}),
 a project of the German state Nordrhein-Westfalen,
 which supported this research in part.} \\
Applied Computer Science VII (AI) \\
University of Hagen \\
58084 Hagen, Germany \\
Sven.Hartrumpf@FernUni-Hagen.De}

\begin{document}

\maketitle

\begin{abstract}
Multiple default inheritance formalisms for lexicons have attracted much
interest in recent years.
I propose a new efficient method to access such lexicons.
After showing two basic strategies for lookup in inheritance lexicons,
a compromise is developed
 which combines to a large degree (from a practical point of view)
 the advantages of both strategies and avoids their disadvantages.
The method is a kind of (off-line) partial evaluation
 that makes a subset of inherited information explicit before using the lexicon.
I identify the parts of a lexicon
 which should be evaluated,
and show how partial evaluation works for inheritance lexicons.
Finally,
the theoretical results are confirmed by a complete implementation.
Speedups by a factor of 10--100 are reached.
\end{abstract}

\section{Introduction}

\subsection{Motivation}
  \label{sec:motivation}

 In recent years,
lexicons based on inheritance formalisms have attracted much interest in
computational linguistics.
These formalisms allow to develop highly structured lexicons with only
 small redundancy.
The property of small redundancy leads to positive consequences, \eg
easy construction, modification, and extension of lexicons and
support for maintaining consistency and conciseness.

Besides the advantages following from nonredundant lexicons,
inheritance lexicons are theoretically attractive
 as they identify linguistically significant classes of lexemes from the vast
 set of all possible classes (cf.~\cite[p.~283]{flickinger_nerbonne92}).
Finally, descriptive parsimony,
 which can be achieved by inheritance lexicons,
is always an advantage.

 Despite the increasing popularity of inheritance lexicons,
lookup for these lexicons is often inefficient,
or at least not as efficient as possible.
There are two basic strategies for access to inheritance lexicons:
\begin{enumerate}
\item Calculate the extension of a lexicon before access,
\ie the extension of each class describing a single lexeme
(a \emw{lexical class}) is completely calculated,
yielding all word forms of this lexeme\footnote{It is assumed that the lexicon
contains morphological information to account for inflection.
If the lexicon describes only base forms and morphology is treated by a
separate component,
a lexeme is linked to only one feature structure,
 which is defined by multiple inheritance.
Besides this difference,
all arguments about the approach presented in this paper remain valid.},
 each with complete lexical information from the class itself and its
direct and indirect superclasses.
This information,
 standardly represented by some kind of feature structure,
is stored in a new expanded lexicon.
The size of the resulting full-form lexicon is maximal;
lookup time is minimal, however,
 because lookup amounts to index search.

Even if a full-form lexicon can be realized for some applications,
it is---from a scientific point of view---still desirable to achieve significant
efficiency improvements
 if they are possible through a reasonable amount of additional work and
 resources.
\item
Calculate an index (or several indices) before access and
look up words by using this index
(cf.\ \cite{russell_ballim_etal92}).\footnote{An extreme access strategy
is to calculate the extension of the lexicon for each lookup.
An anonymous reviewer pointed out to me (among other improvements)
that this strategy is for most applications inadequate.}
For language analysis,
an index from word forms to corresponding lexical classes is needed.
Lookup of a word in such an \emw{indexed inheritance lexicon}
comprises one index search plus extending (normally) one class.
\end{enumerate}

A lookup in an indexed inheritance lexicon is slower than
a lookup in a full-form lexicon,
 because the latter kind of lookup consists only of an index search.
In this paper,
partial evaluation of inheritance lexicons is explained
 which speeds up lookup significantly (compared to indexed inheritance
lexicons).

\subsection{Inheritance-based lexicon formalism}
  \label{sec:ibl}

 In this paper,
the lexicon formalism \emw{IBL} (\emw{inheritance-based lexicon formalism}),
 which \cite{hartrumpf96} has developed and implemented,
is used
 to illustrate the approach presented in this paper.
This formalism is general and powerful and has been successfully applied
 to partially evaluate several lexicons.
The results of this paper can be transferred to other lexicon formalisms
with similar inheritance concepts
(\eg ELU's formalism, \cite{russell_carroll_etal91,russell_ballim_etal92})
quite easily.
IBL's basic concepts are summarized in the following as far as
it is needed to understand partial evaluation for inheritance lexicons.

IBL is heavily inspired by the lexicon formalism of the \quw{Environnement
Linguistique d'Unification} (ELU),
 which uses multiple default inheritance as
 described by \cite{russell_carroll_etal91,russell_ballim_etal92}.
IBL differs from ELU's formalism in four main ways:
feature structures may contain complex disjunctions and complex negations
(not just atomic ones);
predicative constraints can use coroutining to wait for arguments to become
sufficiently instantiated
(for coroutining in logic programming cf.~\cite{clark_mccabe79,sics95});
a class can decide where in the feature structure to inherit information
from a superclass (One may call this \emw{locating inheritance}.);
IBL is strongly typed, \ie all feature structures must have a type,
while types are optional in ELU.

An IBL lexicon consists of type, generator and class definitions.
Types are defined to type feature structures.

IBL's most important concept are \emw{classes}.
A class is defined by its name, a list of direct superclasses
(\emw{superclass list}),
a (possibly empty) \emw{main feature structure} containing definite information,
a (possibly empty) \emw{default feature structure} containing default
information, and
a set of \emw{variant feature structures} defining a set of \emw{variants}.
Variants are mutually exclusive alternatives,
 which can be used to describe different inflectional forms, for instance.
The following class \cow{a},
 which describes basic properties of German adjectives,
exemplifies the use of main, default, and variant feature structures.
\begin{cop}
\item
class a \\
\q main \\
\q\q syn\car cat = a, \\
\q\q concat(mor\car a\_stem, mor\car suffix, form) \\
\q default \\
\q\q lemma = mor\car stem \\
\q variant \\
\q\q syn\car cdegree = pos, \\
\q\q mor\car a\_stem = mor\car pos\_stem \\
\q variant \\
\q\q syn\car cdegree = comp, \\
\q\q mor\car a\_stem = mor\car comp\_stem \\
\q variant \\
\q\q syn\car cdegree = sup, \\
\q\q mor\car a\_stem = mor\car sup\_stem. \\
\end{cop}
This class states in its main feature structure
 that all adjectives belong to category \cow{a} and
 word forms (feature \cow{form}) of adjectives are always the
 concatenation (predicate \cow{concat}) of a specific stem
 \cow{mor\car a\_stem} and the suffix \cow{mor\car suffix}.
 In the next feature structure,
it is expressed
 that the \cow{lemma} is normally equal to the morphological stem.
Finally,
three variants prepare the generation of positive, comparative, and superlative
adjective forms,
 all of which are realized within word boundaries.
The given class is a \emw{nonlexical class}
 as it is not a class
  that provides word forms for a specific lexeme
  (\emw{lexical class}).
This distinction is made by \cite{russell_carroll_etal91}.

The default information coming from a superclass can be overwritten in
subclasses.
Inheritance conflicts are solved by prioritized inheritance
 which uses the concept of \emw{class precedence lists} (CPL)
 as described for the Common Lisp Object System (CLOS) by
 \cite{keene89}.\footnote{The CPL of a class can be derived by
topologically sorting its direct and indirect superclasses,
 so that classes
  that appear together in a superclass list of these superclasses
 keep their order in the CPL
 and subclasses always precede their superclasses.}
The information of a superclass can be inherited by a subclass at a certain
position (\emw{locating inheritance}).
The information inherited can not be restricted or selected
(\cite{hartrumpf96} calls this \emw{selective inheritance}, which is allowed in
DATR \cite{evans_gazdar96}.)
 because this contradicts the principle of data encapsulation and
 may lead to unwieldy and poorly structured lexicons.
To calculate the extension of a lexical class,
one adds to it
the information from main/variant feature structures of its superclasses
by unification and
the information from default feature structures of its superclasses
by default unification.

IBL's third kind of lexical definitions are \emw{generators}.
A generator is a rule
 that can be used to generate a new lexical class from another lexical class.
Such a rule is applied to an element of the extension
 if a special feature \cow{gener} is set to the generator's name.

In the next section,
it is explained
 how a multiple inheritance hierarchy of classes can be partially
 evaluated.

\section{Partial evaluation of inheritance lexicons}

\subsection{Background}

Lexicon lookup in indexed inheritance lexicons
(as described in section~\ref{sec:motivation})
consists of two operations:
first, searching for the relevant class\footnote{To simplify the discussion,
it is assumed that only one class is relevant for a given query.
All arguments remain valid
 if several classes are found in the index.}
in the index;
second, calculating the extension of this class.
As the latter step consumes much more time than the first,
it is the main target for improvements.

The idea is to precompute the extension of lexical classes to reduce
lookup time.
This precomputation represents a kind of (off-line) partial evaluation for
inheritance lexicons (with respect to lexicon lookup).
In general,
there is a trade-off between the size of the partially evaluated lexicon
(\emw{pe-lexicon}) and the work
 that remains to be done at lookup time.
This trade-off can be characterized by the gap between
completely extended inheritance lexicons and indexed inheritance lexicons.

In this paper,
a compromise for this trade-off is proposed
 that yields both:
 speedup of lexicon access by a factor of 10--100
 (compared to indexed inheritance lexicons)
 and moderate lexicon size.

\subsection{Partial evaluation without defaults}
  \label{sec:what_to_evaluate}

For our partial evaluation method,
indexed inheritance lexicons were the starting point.
For indexed inheritance lexicons,
the extension of a lexical class $ c $ is calculated at lookup time according
to algorithm 1 in Figure~\ref{fig:algorithma}.
\begin{figure}
\rule{\columnwidth}{0.1 ex}
The extension $ ext(c) $ of a lexical class $ c $ with CPL
$ cpl(c) = \langle c_1 c_2 \dots c_n \rangle $ is calculated in two steps: \\
Let $ M_i $, $ D_i $, $ V_i $ ($1 \leq i \leq n$) be the main feature
structure, the default feature structure, and the set of variant
feature structures, respectively, of class $ c_i $.
(The big unification operator (\bigunif) is a unification over two sets of
feature structures,
 \ie for two sets $ A $ and $ B $ of feature structures,
$ A \bigunif B $ is defined as follows:
\[
 A \bigunif B := \{ F_A \unif F_B | F_A \in A, F_B \in B \} 
  \setminus \{ \bot \}
\]
$ \dunif $ is a default unification operator.) \\
\begin{eqnarray*}
& 1. & ext_{s}(c) := \{ M_1 \} \bigunif V_1 \bigunif \\
 & & \q \{ M_2 \} \bfbigunif V_2 \bfbigunif \dots \{ M_n \} \bfbigunif V_n \\
& 2. & ext(c) := ( \dots ((ext_{s}(c) \dunif D_1) \dunif D_2) \\
 & & \q \dots \dunif D_n)
\end{eqnarray*}
\caption{Algorithm 1\label{fig:algorithma}}
\rule{\columnwidth}{0.1 ex}
\end{figure}

Step 2 will be discussed in section~\ref{sec:defaults}.
The idea of our approach is to precompute some of the unification
operations in algorithm 1.
Two constraints have to be obeyed:
first, the number and size of \emw{pe-results} (results of partial
evaluation, which (together with the lexical classes) form the pe-lexicon)
must be tolerable;
second, there should remain only little to do during lookup.

If one optimizes in favor of the second constraint by
doing all unification operations for all lexical classes (This is the
complete precomputation strategy.),
the number and size of pe-results is unacceptable.
But if one excludes the lexical class itself from partial evaluation
(\ie only the unification operations printed in boldface ($ \bfbigunif $)
in algorithm 1 are performed during partial evaluation),
the number of pe-results shrinks to a tolerable magnitude,
especially as in a well structured inheritance lexicon the following
holds ($ \mathbf{C_l} $ is the set of all lexical classes in the lexicon):
\begin{eqnarray*}
n_{cpl} := | \{ cpl(c) - c | c \in \mathbf{C_l} \} | \ll \\
\q | \{ cpl(c) | c \in \mathbf{C_l} \} | = | \mathbf{C_l} | =: n_l
\end{eqnarray*}
For the lexicon further discussed in section~\ref{sec:performance},
$ n_l $ is 1510, while $ n_{cpl} $ is 225.

Empirical results (see Table~\ref{tab:runtime_pe} below) confirm
 that the number $ n_{cpl} $ doesn't increase much
 when the number $ n_l $ of lexical classes in the lexicon exceeds a certain
 number,
or to be more specific,
$ n_{cpl}/n_l $ approaches zero for growing $ n_l $.
Therefore, the number of parts to be evaluated will be tolerable,
 if one precomputes only the unification operations
  that are in boldface in algorithm 1.

 At lookup time,
only the information contained in the feature structures of the current lexical
class in the pe-lexicon has to be added by unification to the result of
partial evaluation
(which can be considered to be the only superclass of that lexical class
in the pe-lexicon),
\ie only the unification operations
 that are not in boldface in algorithm 1
have to be calculated.
The number of unifications
 which are precomputed for the extension of a lexical class
is often high (see Table~\ref{tab:runtime_lookup} below)
 as variants may lead to exponential growth of this number.

\subsection{Partial evaluation with defaults}
  \label{sec:defaults}

The approach presented so far is only correct
 if the operations used in calculating extensions are associative
 and commutative
because the normal order of operations (bottom-up) is changed by partial
evaluation.
These conditions are met by the (nondefault) unifications
 used in step 1 of algorithm 1.
 However, to add defaults (cf.\ step 2 of algorithm 1),
normally a default unification operation is used
 which is neither associative nor commutative.
This is also true of the prioritized default unification
 which is used in IBL and which assigns a priority to all atomic
 feature structures of a default feature structure as defined by
 \cite{hartrumpf96}.\footnote{There are associative and commutative default
unification operations
(cf.\ \cite{lascarides_copestake95,lascarides_briscoe_etal96});
 however,
they might be too inefficient for applications
 that heavily use default unification.}

 To solve the correctness problem for defaults,
I use a simple strategy.
They are not treated during partial evaluation.
Instead,
one stores the atomic feature structures of default feature structures
in the pe-lexicon and
combines them with the pe-result (by prioritized default unification) 
at lookup time.

To sum up partial evaluation for inheritance lexicons,
algorithm 1 in Figure~\ref{fig:algorithma} for calculating the
extension of a lexical class
is separated in a partial evaluation part (the algorithm in
Figure~\ref{fig:algorithmb} for calculating a pe-lexicon)
and a lookup part (see next section).

\begin{figure}
\rule{\columnwidth}{0.1 ex}
for all $ c \in \mathbf{C_l} $ do: \\
1.~Let $ cpl(c) = \langle c_1 c_2 \dots c_n \rangle $.
Let $ M_i $, $ D_i $, $ V_i $ ($1 \leq i \leq n$) be the main feature
structure, the default feature structure, and the set of variant
feature structures, respectively, of class $ c_i $. \\
Calculate and store as a pe-result (if not already done for another class
whose CPL differs only in $ c_1 $):
\begin{eqnarray*}
& &\{ M_2 \} \bigunif V_2 \bigunif \dots \{ M_n \} \bigunif V_n
\mbox{ , and} \\
& & \langle D_2 \dots D_n \rangle
\end{eqnarray*}
2.~Store for class $ c $ a reference to this pe-result,
and $ M_1 $, $ V_1 $, $ D_1 $ (the feature structures of class $ c $).
\caption{Algorithm for partial evaluation\label{fig:algorithmb}}
\rule{\columnwidth}{0.1 ex}
\end{figure}

Partial evaluation of inheritance lexicons is exemplified by its effects
on the morphosyntactic description of German adjectives like \quw{klein}
(\quw{small}) in Figure~\ref{fig:pe_example}.
Classes appear below their direct superclasses and are connected by directed
edges.
Class \cow{a} is a graphical representation of the class definition
in section~\ref{sec:ibl}.
Its direct subclass \cow{a\_forms} contains (besides some defaults)
variant feature structures
 that describe the possible suffixes of German adjectives.
A lexical class like \cow{klein} inherits from classes \cow{a\_forms} and
\cow{a} indirectly through the class \cow{a\_decl}
 which contains typical values for the construction of
 predicative forms of adjectives.
The class \cow{klein} itself contains only the idiosyncratic
information
 that the morphological stem for this adjective is \quw{klein}.
The right side of Figure~\ref{fig:pe_example} shows
 that partial evaluation for inheritance lexicons unfolds the lexical hierarchy
 so that the hierarchy becomes flat.
\newsavebox{\classa}
\sbox{\classa}{%
\scriptsize
\begin{tabular}{@{}ll@{}}
 main &
 \begin{avm}
 \[
   form & concat(\avmt{mor\|a\_stem}, \avmt{mor\|suffix}) \\
   syn\|cat & a \\
 \]
 \end{avm}
\\
 default &
 \begin{avm}
 \[
   lemma & \@1 \\
   mor\|stem & \@1 \\
 \]
 \end{avm}
\\
 variant &
 \begin{avm}
 \[
   mor &
   \[
     a\_stem & \@1 \\
     pos\_stem & \@1 \\
   \] \\
   syn\|cdegree & pos \\
 \]
 \end{avm}
\\
 variant &
 \begin{avm}
 \[
   mor &
   \[
     a\_stem & \@1 \\
     comp\_stem & \@1 \\
   \] \\
   syn\|cdegree & comp \\
 \]
 \end{avm}
\\
 variant &
 \begin{avm}
 \[
   mor &
   \[
     a\_stem & \@1 \\
     sup\_stem & \@1 \\
   \] \\
   syn\|cdegree & sup \\
 \]
 \end{avm}
\\
\end{tabular}}
\newsavebox{\classformsb}
\sbox{\classformsb}{%
\scriptsize
\begin{avm}
     \{
       \[
         case & nom $\lor$ acc \\
         gend & fem \\
         decl & 1 \\
       \] \\
       \[
         case & nom \\
         num & sg \\
         decl & 2 \\
       \] \\
       \[
         case & acc \\
         num & sg \\
         gend & $\lnot$masc \\
         decl & 2
       \] \\
       \[
         case & nom $\lor$ acc \\
         num & sg \\
         gend & fem \\
         decl & 3 \\
       \]
     \}
\end{avm}
}
\newsavebox{\classforms}
\sbox{\classforms}{%
\scriptsize
\begin{tabular}{@{}ll@{}}
 default & \% \coc{d$_{a\_forms}$ omitted}
\\
 variant &
 \begin{avm}
 \[
   \avmspan{mor\|suffix \avmvalue{\dq e\dq}} \\
   syn &
   \[
     agr & \usebox{\classformsb} \\
     use & attr \\
   \]
 \]
 \end{avm}
\\
 variant & \% \coc{variant for attributive suffix \quw{-em} omitted} \\
 variant & \% \coc{variant for attributive suffix \quw{-en} omitted} \\
 variant & \% \coc{variant for attributive suffix \quw{-er} omitted} \\
 variant & \% \coc{variant for attributive suffix \quw{-es} omitted} \\
 variant & \% \coc{variants for predicative suffixes omitted} \\
\end{tabular}
}
\newsavebox{\classdecl}
\sbox{\classdecl}{%
\scriptsize
\begin{tabular}{@{}ll@{}}
 main &
 \begin{avm}
 \[
   mor &
   \[
     pred\_0 & + \\
     pred\_e & - \\
   \] \\
 \]
 \end{avm}
\\
\end{tabular}
}
\begin{figure*}[thp]
\scriptsize
\begin{center}
\textbf{\normalsize Classes}
\vspace*{2 ex}

\mbox{\NoPSframes\xymatrix@R=6.0ex{
*+[F]\txt{\parbox{26 em}{
class {\bf a} \\
\usebox{\classa}
}} &
{}\save[]+<0cm,-34ex>*+[F]\txt{\parbox{26 em}{
{\bf pe-result 71} \% \coc{corresponds to $cpl(\mbox{\bf klein}) - \mbox{\bf klein} $ } \\
\% \coc{$ = \langle \mbox{\bf a\_decl } \mbox{\bf a\_forms } \mbox{\bf a}
  \rangle $} \\
1: \% \coc{positive form with suffix \quw{-e}} \\
 \begin{avm}
 \[
   form & concat(\avmt{mor\|a\_stem}, \avmt{mor\|suffix}) \coc{(delayed)} \\
   mor &
   \[
     a\_stem & \@1 \\
     pos\_stem & \@1 \\
     pred\_0 & + \\
     pred\_e & - \\
     suffix & \dq e\dq \\
   \] \\
   syn &
   \[
     agr & \usebox{\classformsb} \\
     cat & a \\
     cdegree & pos \\
     use & attr \\
   \]
 \]
 \end{avm} \\
2: \% \coc{positive form with suffix \quw{-em} omitted} \\
\vdots \\
6: \% \coc{predicative positive form with suffix \quw{-$\epsilon$} omitted} \\
7: \% \coc{comparative form with suffix \quw{-e} omitted} \\
\vdots \\
18: \% \coc{predicative superlative form with suffix \quw{-sten} omitted} \\
defaults
 \begin{avm}
 \<
   \avmvalue{d$_{a\_forms}$},
   \[
     lemma & \@1 \\
     mor\|stem & \@1 \\
   \]
 \>
 \end{avm}
}}
\ar@{<-}[]!<0cm,-88.5ex>;[ddd]+UC\restore
\\
*+[F]\txt{\parbox{26 em}{
class {\bf a\_forms} \\
\usebox{\classforms}
}} \ar[u]
\\
*+[F]\txt{\parbox{26 em}{
class {\bf a\_decl} \\
\usebox{\classdecl}
}} \ar[u]
\\
*+[F]\txt{\parbox{26 em}{
lexical class {\bf klein (ID 466)} \\
\begin{tabular}{@{}ll@{}}
 main &
 \begin{avm}
 \[
   mor\|stem & \dq klein\dq \\
 \]
 \end{avm}
\\
\end{tabular}
}} \ar[u]
&
*+[F]\txt{\parbox{26 em}{
lexical class {\bf klein (ID 466)} \\
\begin{tabular}{@{}ll@{}}
 main &
 \begin{avm}
 \[
   mor\|stem & \dq klein\dq \\
 \]
 \end{avm}
\\
\end{tabular}
}} 
}}

\vspace*{6 ex}
\textbf{\normalsize Index}

\begin{tabular}{p{28 em}p{28 em}}
\begin{center}
\begin{tabular}{lll}
klein     & $\mapsto$ & 466 \\
kleine    & $\mapsto$ & 466 \\
kleinem   & $\mapsto$ & 466 \\
kleinen   & $\mapsto$ & 466 \\
kleiner   & $\mapsto$ & 466 \\
kleinere  & $\mapsto$ & 466 \\
kleinerem & $\mapsto$ & 466 \\
kleineren & $\mapsto$ & 466 \\
kleinerer & $\mapsto$ & 466 \\
kleineres & $\mapsto$ & 466 \\
kleines   & $\mapsto$ & 466 \\
kleinste  & $\mapsto$ & 466 \\
kleinstem & $\mapsto$ & 466 \\
kleinsten & $\mapsto$ & 466 \\
kleinster & $\mapsto$ & 466 \\
kleinstes & $\mapsto$ & 466 \\
\end{tabular}
\end{center}
&
\begin{center}
\begin{tabular}{lll}
klein     & $\mapsto$ & 466,\{6\} \\
kleine    & $\mapsto$ & 466,\{1\} \\
kleinem   & $\mapsto$ & 466,\{2\} \\
kleinen   & $\mapsto$ & 466,\{3\} \\
kleiner   & $\mapsto$ & 466,\{4,12\} \\
kleinere  & $\mapsto$ & 466,\{7\} \\
kleinerem & $\mapsto$ & 466,\{8\} \\
kleineren & $\mapsto$ & 466,\{9\} \\
kleinerer & $\mapsto$ & 466,\{10\} \\
kleineres & $\mapsto$ & 466,\{11\} \\
kleines   & $\mapsto$ & 466,\{5\} \\
kleinste  & $\mapsto$ & 466,\{13\} \\
kleinstem & $\mapsto$ & 466,\{14\} \\
kleinsten & $\mapsto$ & 466,\{15,18\} \\
kleinster & $\mapsto$ & 466,\{16\} \\
kleinstes & $\mapsto$ & 466,\{17\} \\
\end{tabular}
\end{center}
\end{tabular}
\end{center}
\caption{Effects of partial evaluation for the German adjective
  \quw{klein}\label{fig:pe_example} (left side: original lexicon; right side:
  pe-lexicon)}
\end{figure*}

\subsection{Partial evaluation and lexical rules}
  \label{sec:rules}

Lexical rules are considered important in many lexicon formalisms.
Although some lexical rules can be replaced by using the variant concept
(cf.\ section~\ref{sec:ibl}),
there remain cases
 where lexical rules are useful, \eg for derivational morphology.
The role of lexical rules is played in IBL by generators (cf.\ section~\ref{sec:ibl}).
A generator consists of two parts:
a superclass list and
a mapping from the current feature structure (the input feature structure)
to a new feature structure (the output feature structure).
A straightforward treatment of generators by partial evaluation is
 to store all these output feature structures as new lexical classes
 (containing only a main feature structure)
 in the pe-lexicon and to treat the CPL corresponding to the
 superclass list of a generator like a CPL of a lexical class.\footnote{In
 some systems,
it can be useful to delay the application of certain lexical rules.}

\section{Access to partially evaluated lexicons}

A pe-lexicon consists of lexical classes and pe-results.
Each lexical class refers to one pe-result
 that corresponds to its CPL (without the lexical class itself).
A pe-result $ p $ comprises a list $ p_f $ of feature structures and
a list $ p_d $ of default atomic feature structures
(cf.\ upper right part of Figure~\ref{fig:pe_example}).

 To access a pe-lexicon,
an index is used
 which is identical to the one used for the original indexed inheritance
 lexicon,
 \ie a mapping from a set of key features to the classes
  whose extensions contain relevant information.
 For language analysis,
this set of key features usually consists only of the
(orthographical or phonological) form feature;
such an index is assumed in the following.
After searching for the relevant lexical class $ c $
in the index\footnote{Again, I simplify
by describing only cases with one relevant class.},
the extension of class $ c $ has to be determined.
First, the pe-result $ p_f $ for class $ c $ is retrieved from the pe-lexicon.
Second, the main and variant feature structures of class $ c $ are added by
unification.
Third, the default atomic feature structures of class $ c $ and the default
atomic feature structures in $ p_d $ are concatenated and added by default
unification.
These steps lead to the (complete) extension of class $ c $.

The approach described is slower than necessary
 when $ p_f $ contains many feature structures (as it is often the case for
 highly inflected languages)
 because the second and third step must be done for every member of $ p_f $.
This situation is substantially improved
by considering only those elements of $ p_f $
 that are relevant for the word form $ w $ of a given query.
 In order to follow this strategy,
the index is extended.
A word form index has the signature
$ \mathbf{W} \rightarrow \mathbf{C_l} \times 2^{\mathbb{N}} $
(instead of just $ \mathbf{W} \rightarrow \mathbf{C_l} $; $ \mathbf{W} $ is the
set of all word forms, $ \mathbf{C_l} $ is the set of all lexical classes in
the lexicon).
The additional set $ S $ of natural numbers in an index entry
$ w \mapsto (c, S) $ identifies those elements of $ p_f $
(of the pe-result for $ cpl(c) - c $)
 that are relevant for the given word form $ w $.
Figure~\ref{fig:pe_example} shows a part of an index for
an indexed inheritance lexicon and the corresponding part of an index for a
pe-lexicon,
and Figure~\ref{fig:algorithmc} contains the complete algorithm for lookup in
pe-lexicons.

\begin{figure}
\rule{\columnwidth}{0.1 ex}
1.~$ w \mapsto (c, S) $ = index entry for word form $ w $ \\
(Let the lexical class $ c $ refer to a pe-result $ p $ consisting of
feature structures $ p_f $ and atomic default feature structures
$ p_d = \langle D_1 D_2 \dots D_n \rangle $.) \\
2.~Calculate for the feature structures $ PE $ corresponding to
feature structures $ S $ in $ p_f $ the following
($ M_c $, $ D_c $, $ V_c $ are the main feature structure,
the default feature structure, and the variant feature structures for
the lexical class $ c $, respectively): \\
\begin{eqnarray*}
& 1. & ext'_s(c) := \{ M_c \} \bigunif V_c \bigunif PE \\
& 2. & ext'(c)   :=
  ( \dots (((ext'_{s}(c) \dunif D_c) \dunif D_1) \dunif \\
 & & \q D_2) \dots \dunif D_n)
\end{eqnarray*}
\caption{Algorithm for lookup in pe-lexicons\label{fig:algorithmc}}
\rule{\columnwidth}{0.1 ex}
\end{figure}

\section{Performance of partially evaluated lexicons}
  \label{sec:performance}

\subsection{Performance of partial evaluation}

As partial evaluation is done in a separate compilation step,
the size of the resulting pe-lexicon is much more important
than the runtime (which is linear in the number $ n_l $ of lexical
classes).
A pe-lexicon\footnote{In order not to restrict the generality of the
empirical results in section~\ref{sec:performance},
the specific treatment of generators (cf.\ section~\ref{sec:rules}) is
excluded here.}
contains in addition to the lexical classes of the original
lexicon (with superclass lists replaced by references to pe-results,
cf.\ section~\ref{sec:what_to_evaluate})
a set of pe-results.
If one follows the strategy described in section~\ref{sec:what_to_evaluate}
the size of the pe-lexicon grows only linearly to $ n_{cpl} $.

The number $ n_{cpl} $ is bounded by the number $ n_l $ of lexical classes;
in realistic lexicons,
$ n_{cpl} $ is much smaller than $ n_l $,
 as explained in section~\ref{sec:what_to_evaluate}.
Therefore,
the size increase for pe-lexicons is tolerable
 as reflected in Table~\ref{tab:runtime_pe}.
The lexicons of this table were created by adding lexemes in the order of
a frequency list (see \cite[pp.~165--219]{rosengren77}),
 which is based on a German newspaper corpus.
\begin{table*}[tbh]
\begin{center}
\begin{tabular}{@{}rrrrrrrc@{}}
\toprule
\multicolumn{4}{c}{Lexicon} & \multicolumn{3}{c}{pe-Lexicon} & Size Increase \\
\cmidrule(r){1-4}\cmidrule(lr){5-7}\cmidrule(l){8-8}
$ n_l $ & $ n_n $ & $ n_{cpl} $ & $ n_{lfs} $ &
  $ t_e $ (s) & $ t_e / n_{l} $ (s) & $ n_{lfs} $ &
  $\frac{n_{lfs}(pe-lexicon)}{n_{lfs}(lexicon)}$ \\
\midrule
 336 &  96 & 107 &  724 &  24.8 & 0.074 & 1753 & 2.42 \\
 755 & 118 & 158 & 1196 &  48.2 & 0.064 & 3005 & 2.51 \\
1510 & 145 & 225 & 2003 &  94.3 & 0.062 & 5193 & 2.59 \\
\bottomrule
\end{tabular}
\vspace*{2 ex}

\textbf{Legend}
\vspace*{2 ex}

\begin{tabular}{@{}ll@{}}
$ n_l $ & number of lexical classes \\
$ n_n $ & number of nonlexical classes \\
$ n_{cpl} $ & number of different CPLs of lexical classes (excluding the
lexical class itself) \\
$ n_{lfs} $ & number of feature structures in the lexicon \\
$ t_e $ & evaluation time for producing the pe-lexicon (SICStus Prolog 3 on a SUN Ultra 1) \\
\end{tabular}
\end{center}
\caption{Experimental results for partial evaluation\label{tab:runtime_pe}}
\end{table*}

\subsection{Performance of lexicon lookup}

A lookup in a pe-lexicon comprises two steps:
an index entry is searched and
a stored partial extension is completed.
The first step can be done by standard methods from computer science,
\eg by using hash tables which allow constant time search.

During the second step,
one main feature structure and several variant feature structures have to
be added by unification and
the defaults of the pe-result and of the lexical class have to be
added by default unification (as specified in Figure~\ref{fig:algorithmc}).
In addition,
there might be predicative constraints
 that had to be delayed during partial evaluation.
The coroutining mechanism will initiate their evaluation during this step.
This is the worst case;
a typical lexical class comprises only a small main feature structure,
 if the lexicon is well structured.\footnote{If the feature structure
 that is to be added by unification
contains only few atomic feature structures,
the operation is sped up,
as the operation used need not be (and is not) a full unification:
the atomic feature structures only have to constrain the current
intermediate result of calculating the extension of a class;
the other way of constraining---as it is required by full unification---is
not necessary.}

The results in Table~\ref{tab:runtime_lookup} show
 that partial evaluation achieves significant speedups,
 especially for complex lexical classes.
On the average,
lookup in a pe-lexicon is 60 times faster than in the original indexed
inheritance lexicon.
\begin{table*}[tbh]
\begin{center}
\begin{tabular}{@{}lrrrrrrrrrc@{}}
\toprule
Word Form & \multicolumn{4}{c}{Lexicon} &
  \multicolumn{4}{c}{pe-Lexicon} & Speedup \\
\cmidrule(r){2-5}\cmidrule(lr){6-9}\cmidrule(l){10-10} &
  $ n_s $ & $ n_{fs} $ & $ n_{afs} $ & $ t_l $ (ms) &
  $ n_{fs} $ & $ n_{afs} $ & $ n_{dp} $ & $ t_l $ (ms) &
  $\frac{t_{l}(lexicon)}{t_{l}(pe-lexicon)}$ \\
\midrule
\quw{wegen} (preposition)    & 1    &   4 &    7 &   8.0 & 2 &  3    & 0    & 0.7 & 11.4 \\
\quw{Kreises} (noun)         & 2    &  18 &   52 &  31.6 & 2 &  8    & 0    & 1.2 & 26.3 \\
\quw{sch\"onste} (adjective) & 4    &  72 &  332 & 168.8 & 2 &  9    & 1    & 4.9 & 34.4 \\
\quw{laufe} (verb)           & 7    & 538 & 2204 & 385.1 & 2 & 24    & 2    & 5.3 & 72.7 \\
average (1510)
               & 3.63 & 153 &  631 & 133.2 & 2.79 & 10.06 & 0.44 & 2.2 & 60.5 \\
\bottomrule
\end{tabular}
\vspace*{2 ex}

\textbf{Legend}
\vspace*{2 ex}

\begin{tabular}{@{}ll@{}}
$ n_s $ & number of superclasses \\
$ n_{fs} $ &
  number of feature structures in superclasses (added during extending) \\
$ n_{afs} $ &
  number of atomic feature structures in superclasses (added during extending) \\
$ t_l $ & lookup time (SICStus Prolog 3 on a SUN Ultra 1) \\
$ n_{dp} $ & number of delayed predicates \\
\end{tabular}
\end{center}
\caption{Experimental results for lexicon lookup\label{tab:runtime_lookup}}
\end{table*}

\section{Related work}

Partial evaluation is a technique that has been
applied---in different ways---in several areas:
in logical, functional, and imperative programming
(see for instance \cite{jones_gomard_etal93} for a thorough introduction and
\cite{danvy_gluck_etal96} for current research);
for efficient access to methods in programming languages with
single inheritance (\cite{khoo_sundaresh91}); etc.
The problem of partial evaluation for inheritance lexicons differs in two ways:
while behavior is limited to lookup functions,
the data is complex due to its large size and high expressivity
(multiple inheritance, defaults, disjunctions, negations).

In the context of inheritance lexicons,
\cite[p.~240]{copestake93} mentions for the LKB
(lexical knowledge base, \cite{copestake93b})
 that expanded classes (\quw{expanded psorts} in LKB) \quw{can be cached},
 but does not provide any details.
 If inheritance proceeds top-down,
  as in the LKB,
partial evaluation will be somewhat easier
 because the information
  that is added from the lexical class
 has always lower priority than the information of its superclasses
 and thus defaults don't have to be treated specially by partial evaluation.
 However,
this practice is for most applications counter-intuitive (although it is
justified for the original domain of LKB,
semiautomatic extraction of lexical information from machine-readable
dictionaries,
as \cite[p.~239]{copestake93} shows)
 because normally a class should have precedence over its superclasses.

\section{Perspectives}

The results for partial evaluation of inheritance lexicons
presented in this paper improve the practical use of such lexicons
significantly.
Interesting questions for further research remain, however.

The number of pe-results might be too high for some combinations
of lexicons and applications.
 To reduce this number,
the set of classes can be partitioned.
 According to these partitions,
the CPL of a class is split up in parts and
only those partial CPLs are evaluated.
During lexicon lookup,
it will be necessary to combine results from different partial CPLs.
This combination will be easier if the class partitions are orthogonal.
For example,
if, in the lexicon at hand, morphosyntactic and semantic information don't
interact much,
one may use a class partition \cow{mor\_syn} and a class partition \cow{sem}.

Two more questions arise from the first:
How can useful partitions of classes be linguistically defined or
automatically determined?
And: How do these partitions influence lookup time and size of pe-lexicons?

Another perspective is the use of results from partial evaluation
 to infer new linguistically relevant classes in inheritance lexicons
 semi-automatically.

\nocite{briscoe_copestake_etal93}
\begin{scriptsize}

\end{scriptsize}

\end{document}